\documentclass{article}
\usepackage{spconf,amsmath,graphicx}
\usepackage{soul}

\usepackage{enumitem}
\setlist{nosep, leftmargin=14pt}

\usepackage{mwe} 
\usepackage{multirow}
\usepackage{url}

\title{A NEW LOGIC FOR PEDIATRIC BRAIN TUMOR SEGMENTATION}

\name{\begin{tabular}{c}Max Bengtsson $^{1}$\!\! \quad Elif Keles  $^{1}$\!\! \quad Gorkem Durak $^{1}$\!\! \quad Syed Anwar  $^{2}$\!\! \\ \quad \textit{Yuri S. Velichko} ${^1}$\!\!  \quad \textit{Marius G. Linguraru} $^{2}$\!\!  \quad \textit{Angela  J. Waanders} $^{1,3}$\!\! \quad \textit{Ulas Bagci} $^{1}$\end{tabular}}
\address{ \small{$^{1}$Northwestern University, Feinberg School of Medicine, Chicago IL, USA} \\
\small{$^{2}$Children’s National Hospital, Sheikh Zayed Institute for Pediatric Surgical Innovation, Washington, DC, USA} \\
 \small{$^{3}$ Ann \& Robert H Lurie Children's Hospital of Chicago, Chicago, IL, USA} 
 }
\begin{document}
\small
%
\maketitle
\begin{abstract}
\label{sec:abstract}
In this paper, we present a novel approach for segmenting pediatric brain tumors using a deep learning architecture, inspired by expert radiologists' segmentation strategies. Our model delineates four distinct tumor labels and is benchmarked on a held-out PED BraTS 2024 test set (i.e., pediatric brain tumor datasets introduced by BraTS). Furthermore, we evaluate our model's performance against the state-of-the-art (SOTA) model using a new external dataset of 30 patients from CBTN (Children's Brain Tumor Network), labeled in accordance with the PED BraTS 2024 guidelines and 2023 BraTS Adult Glioma dataset. We compare segmentation outcomes with the winning algorithm from the PED BraTS 2023 challenge as the SOTA model. Our proposed algorithm achieved an average Dice score of 0.642 and an HD95 of 73.0 mm on the CBTN test data, outperforming the SOTA model, which achieved a Dice score of 0.626 and an HD95 of 84.0 mm. Moreover, our model exhibits strong generalizability, attaining a 0.877 Dice score in whole tumor segmentation on the BraTS 2023 Adult Glioma dataset, surpassing existing SOTA. Our results indicate that the proposed model is a step towards providing more accurate segmentation for pediatric brain tumors, which is essential for evaluating therapy response and monitoring patient progress. Our source code is available at \url{https://github.com/NUBagciLab/Pediatric-Brain-Tumor-Segmentation-Model}.

\end{abstract}
\section{Introduction}
\label{sec:intro}
Automated brain tumor segmentation provides precise and objective quantification, enhancing the accuracy and consistency of tumor burden assessment and treatment response monitoring, which could significantly improve disease prognosis and support personalized treatment planning. This is valuable for patients irrespective of age. However, pediatric brain tumor segmentation presents unique challenges compared to adult cases
due to several factors~\cite{kazerooni2023brain}: anatomical differences, tumor heterogeneity, limited data, and ethical considerations. Additionally, pediatric brain tumors vary significantly between age groups within the pediatric population. 
The constant development and plasticity of the pediatric brain introduces further complexity, as evolving brain structures may impact tumor appearance and growth patterns. Addressing these challenges requires the development of specialized segmentation algorithms that can effectively handle the unique characteristics associated with pediatric brain tumors. 


Towards the dataset problem, over the past two years, the Brain Tumor Segmentation (BraTS) challenge, a community based initiative to develop machine learning models for automating
brain tumor segmentation, introduced the pediatric (PED) segmentation task to establish benchmark methods in this area. Moreover, Children’s Brain Tumor Network
(CBTN), established in 2011, is a collaborative consortium with a data repository for pediatric brain tumor research.
CBTN and PED BraTS initiated efforts to drive research development, with the introduction of pediatric brain
tumor segmentation task at BraTS 2023 \cite{kazerooni2023brain,lilly2023children}. We leverage these data sources in this study, and propose a novel deep-learning segmentation architecture incorporating simple radiological reasoning to label tumor components more accurately. 

Our contributions are summarized as follows:

\begin{figure*}[t]
    \centering
    \includegraphics[width=0.9\textwidth]{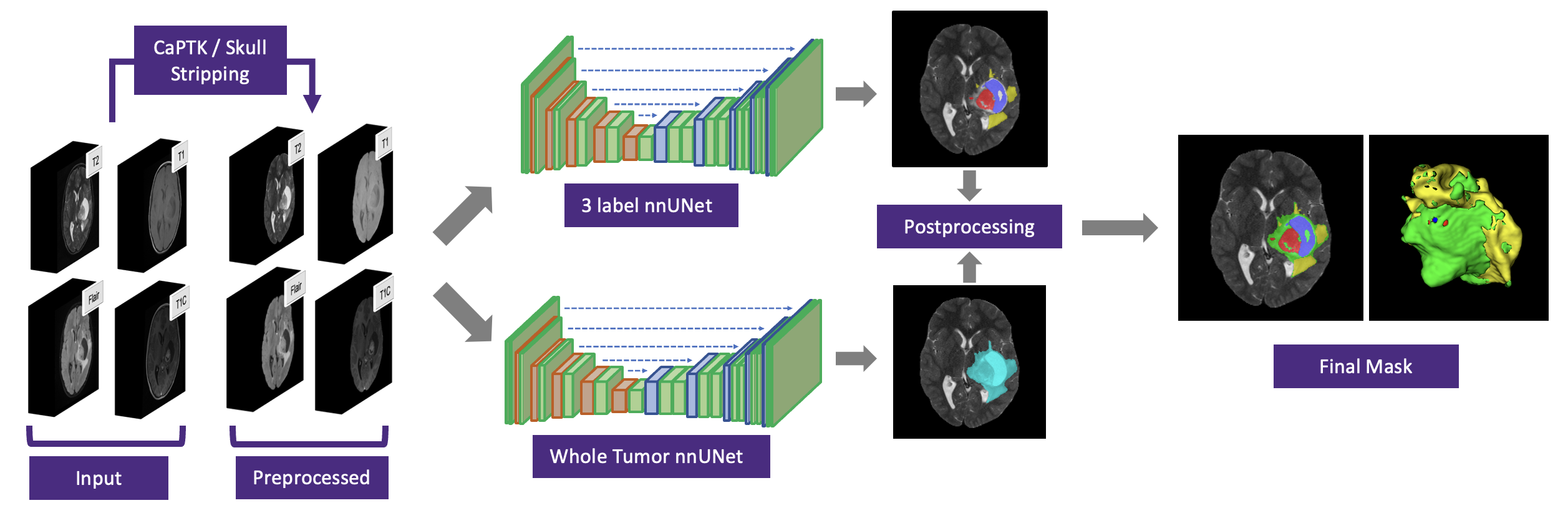}
    \caption{\small{Our model architecture utilizing two nnU-Net frameworks, one trained on ET, CC and ED detection and the other trained to detect WT.}}
    \label{fig:arch}
\end{figure*}

\begin{enumerate}
    \item  We developed a novel yet simple deep learning architecture, inspired by radiological reasoning for 3D multi-class delineation of brain tumors, to segment pediatric brain tumors and their counterparts. 
    \item We evaluated the robustness of our proposed architecture across multiple frameworks, demonstrating significant improvements over existing baseline models.
    \item Our model outperformed the state-of-the-art (SOTA) model—the winner of the PED BraTS 2023 challenge— when tested on an out-of-distribution, real-world dataset from CBTN and BraTS 2023 Adult Glioma dataset.
\end{enumerate}

\section{Methodology}
\label{sec:methods}

\subsection{Dataset and Pre-processing}
\label{ssec:preproc}
\textbf{PED BraTS 2024:} We used the PED BraTS 2024 pediatric challenge dataset~\cite{kazerooni2024brain}, which comprises 261 patients, each with 4 MRI sequences: T1-weighted (T1W), T1 contrast-enhanced (T1CE), T2 weighted (T2W), and T2 FLAIR (T2FL). We held 26 patients as a test set to evaluate the performance of different architectures, namely SegMamba and nnU-Net, and to compare baseline labeling against a 3 Label/Whole Tumor (3L/WT) labeling approach that we propose herein. Tumors were annotated into sub-regions including enhanced tumor (ET), non-enhanced tumor (NET), cystic component (CC), and edema (ED) \cite{kazerooni2024brain}. From these annotations, tumor core (TC) and whole tumor (WT) labels were derived by combining ET, NET, and CC, and all four labels, respectively \cite{kazerooni2024brain}. The PED BraTS 2024 dataset underwent preprocessing using the CaPTK BraTS pipeline, which included reorientation to LPS/RAI (Left, Posterior, Superior / Right, Anterior, Inferior) and rigid image registration to the SRI-24 Atlas \cite{pati2020cancer,davatzikos2018cancer}. Utilizing the already preprocessed data, we skull-stripped the dataset using a deep learning algorithm \cite{vossough2024training}.

\textbf{CBTN Data:} When evaluating performance against the SOTA, we inferred on a testing set of 30 LGG preoperative patients’ MRIs from the CBTN dataset (T1W, T1CE, T2W and T2FL). We used the same CaPTK preprocessing pipeline and skull stripping as in training. 
All sequences were resized to $240\times240\times155$. Two expert radiologists manually segmented the 30 cases from CBTN adhering to the 4 tumor subregion labeling scheme set forth by PED BraTS 2024. This resulted in labels including ET, NET, CC, and ED.

\textbf{BraTS 2023 Adult Glioma Data}: When evaluating our models' reproducibility and generalizability in other datasets, we used the BraTS 2023 Adult Glioma dataset \cite{baid2021rsna}, including T1W, T1CE, T2W, and T2FL. From the training partition of the dataset, we tested using 100 patients annotated with ET, NCR (Necrotic Core) and ED.

\begin{figure*}[t]
    \centering
    \includegraphics[width=0.95\linewidth]{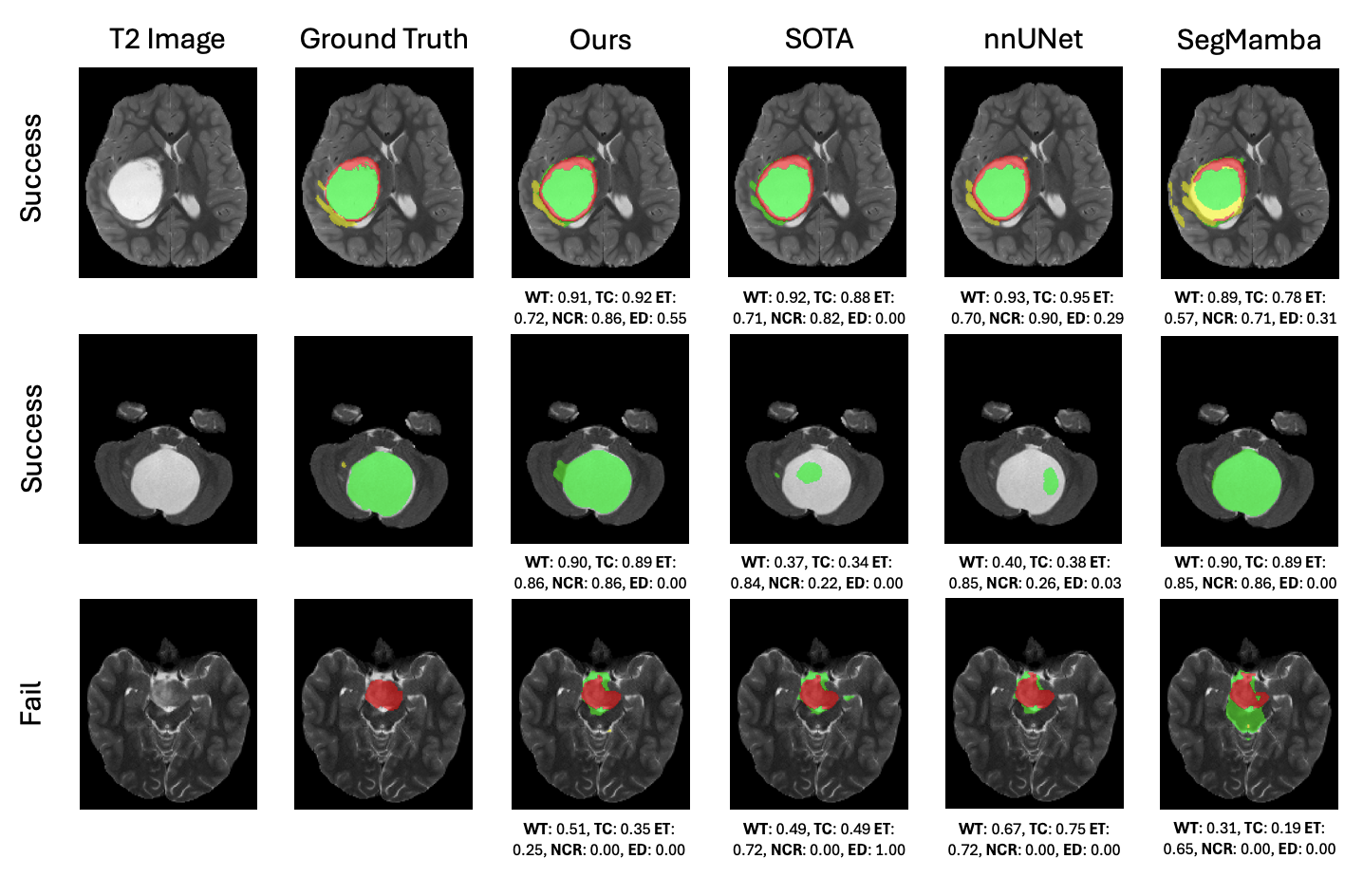} 
    \caption{\small{Example delineations  from  proposed model against SOTA (lesion-wise Dice scores are shown for entire volumetric segmentation).}}
    \label{fig:segs}
\end{figure*}

{
\setlength{\tabcolsep}{3pt}
\begin{table*}[htbp]
    \centering
    \small
    \caption{\small{Lesion-wise Dice Scores, HD95 (mm), Precision and Recall for different 3L/WT models vs. baselines in BraTS 2024 testing set.}}
    \begin{tabular}{|c|cccccc|cccccc|cc|cc|}
        \hline
        \multirow{2}{*}{} & \multicolumn{6}{c|}{\textbf{Lesion-wise Dice Scores}} & \multicolumn{6}{c|}{\textbf{Lesion-wise HD95 (mm)}} & \multicolumn{2}{c|}{\textbf{Precision}} & \multicolumn{2}{c|}{\textbf{Recall}}\\
        \cline{2-17}
        Model & WT & TC & ET & NET & CC & ED & WT & TC & ET & NET & CC & ED & WT & TC & WT & TC\\
        \hline
         SegMamba \cite{xing2024segmamba} & 0.827 & 0.815 & 0.622 & 0.773 & 0.789 & 0.770 & 26.04 & 26.86 & 102.77 & 20.17 & 47.65 & 51.65 & 0.902 & 0.896 & 0.846 & 0.860\\
         SegMamba 3L/WT & 0.850 & 0.832 & 0.622 & 0.790 & 0.813 & 0.788 & 18.38 & 19.47 & 82.79 & 12.87 & 46.98 & 59.28 & 0.922 & 0.896 & 0.904 & 0.908\\
         nnU-Net \cite{isensee2021nnu} & 0.844 & 0.832 & 0.690 & 0.791 & \textbf{0.838} & 0.782 & 25.79 & 26.34 & \textbf{60.64} & 19.92 & \textbf{14.22} & 66.12 & 0.913 & 0.909 & 0.872 & 0.884\\
         nnU-Net 3L/WT & \textbf{0.866} & \textbf{0.853} & \textbf{0.691} & \textbf{0.813} & 0.829 & \textbf{0.855} & \textbf{11.67} & \textbf{12.36} & 61.08 & \textbf{5.44} & 32.69 & \textbf{41.55} & \textbf{0.923} & \textbf{0.915} & \textbf{0.915} & \textbf{0.930}\\
        \hline
    \end{tabular}
    \label{tab:bratsScores}
\end{table*}
}

\subsection{Proposed Model Architecture: 3L/WT}
\label{ssec:arch}
Our model architecture combines two different nnU-Net~\cite{isensee2021nnu} frameworks which are trained to classify different parts of the tumor and combined during post processing to increase segmentation accuracy. Let \(\mathcal{M}_{WT}\) and \(\mathcal{M}_{ET,CC,ED}\) represent these two neural networks based on the nnU-Net framework:
\begin{itemize}
    \item \(\mathcal{M}_{WT}\): A model trained to classify the WT by combining the ET, NET, CC, and ED into a single WT label.
    \item \(\mathcal{M}_{ET,CC,ED}\): A model trained specifically to distinguish between ET, CC, and ED regions within the tumor.
\end{itemize}
The final model output \(\mathcal{S}_{final}\) is derived by combining outputs of \(\mathcal{M}_{WT}\) and \(\mathcal{M}_{ET,CC,ED}\) through a post-processing overlay procedure. Let \(\mathcal{M}_{ET,CC,ED}(x)\) overlay onto \(y_{WT}\) such that regions covered by \(y_{ET}\), \(y_{CC}\), and \(y_{ED}\) are excluded. Here, \(y_{WT}\) is a binary mask for WT, while \(y_{ET}\), \(y_{CC}\), and \(y_{ED}\) represent segmentation masks for ET, CC, and ED regions, respectively. The residual regions within \(y_{WT}\) not covered by \(\{y_{ET}, y_{CC}, y_{ED}\}\) are classified as NET:
\begin{equation}
    y_{NET} = y_{WT} \setminus (y_{ET} \cup y_{CC} \cup y_{ED}).
\end{equation}
This yields the final segmentation mask:
\begin{equation}
    \mathcal{S}_{final} = \{ y_{ET}, y_{CC}, y_{ED}, y_{NET} \}.
\end{equation}

Our rationale for adopting this approach is consistent with how radiologists categorized the labels during manual segmentation. ET, CC, and ED were sub-regions that had specific traits for radiologists to find in order to classify them into their respective classes. However, for NET subregion, this was categorized as any part of the tumor that did not fall into the other three categories \cite{kazerooni2024brain,kazerooni2024brats}. Therefore, we utilized the same approach for the model output. We first classified the WT, find the three labels with clearly discernible differences, and then labeled the remaining region as NET \cite{kazerooni2024brain}. 

For robustness testing, we implemented the above \(3L/WT\) labeling approach using the newly popular SegMamba backbone \cite{xing2024segmamba}. We compared its performance against an ET, NET, CC, and ED trained model to other backbones while taking advantage of SegMamba's global perception.

\subsection{Training and Performance Metrics}
\label{ssec:training}
Model training was performed using 3 Nvidia A6000 GPUs (48 GB) for all nnU-Net frameworks. All nnU-Net models including baseline (4 label), 3 label, and whole tumor models were all trained using 5 fold cross validation and 1000 epochs per fold. The SegMamba models were trained on one Nvidia A6000 GPU (48 GB). Our baseline model (4 label) was trained without folds for 2000 epochs. Both the 3 label and whole tumor models were trained without folds for 1000 epochs. All SegMamba models were trained using the AdamW optimizer with a learning rate 0.00001 and a cosine annealing scheduler with linear warm up for the first 10\% of training. A combination of Dice and cross-entropy loss was used. Both SegMamba and nnU-Net used patch sizes of $128\times128\times128$ and a sliding window inferer during inference. Preprocessing using the CaPTK pipeline took approximately 50 seconds per patient using an Intel Core i9-10900X and skull stripping on a single RTX A6000 took 30 seconds per patient.

For evaluating the performance of our model we adhered to the metrics set forth by PED BraTS 2024 \cite{kazerooni2024brain}. This challenge included unique metrics that calculated both Dice scores and Hausdorff 95 (HD95) distances on a lesion-wise basis \cite{kazerooni2024brain}. It is worth noting that our model segments four labels (ET, NET, CC, ED) but the SOTA does not. The SOTA segments ET, ED and the non-enhancing component (NC) which is composed of NET, CC, and necrosis \cite{capellan2024model}. Therefore, through combining NET and CC, we were able to make a direct comparison using the PED BraTS 2024 metric system. This same process was used for the adult dataset since the ground truths are labeled as ET, NC, and ED. When comparing 3L/WT models on PED BraTS 2024 testing set against baselines, the original labeling system was used without conversion to NC.

{
\setlength{\tabcolsep}{3pt}
\begin{table*}[htbp]
    \centering
    \small
    \caption{\small{Lesion-wise Dice Scores, HD95 (mm), Precision and Recall comparing our model against the SOTA in CBTN testing set.}}
    \begin{tabular}{|c|cccccc|cccccc|cc|cc|}
        \hline
        \multirow{2}{*}{} & \multicolumn{6}{c|}{\textbf{Lesion-wise Dice Scores}} & \multicolumn{6}{c|}{\textbf{Lesion-wise HD95 (mm)}} & \multicolumn{2}{c|}{\textbf{Precision}} & \multicolumn{2}{c|}{\textbf{Recall}}\\
        \cline{2-17}
        Model & WT & TC & ET & NC & ED & AVG & WT & TC & ET & NC & ED & AVG & WT & TC & WT & TC\\
        \hline
         nnU-Net \cite{isensee2021nnu} & 0.621 & 0.644 & 0.525 & 0.466 & 0.518 & 0.555 & 59.09 & \textbf{39.13} & 145.66 & 71.28 & 161.41 & 95.31 & \textbf{0.715} & 0.698 & 0.720 & 0.708\\
         
         SegMamba 3L/WT & 0.601 & 0.585 & 0.451 & 0.447 & 0.656 & 0.548 & 77.11 & 75.23 & 176.69 & 98.47 & 109.68 & 107.44 & 0.652 & 0.639 & 0.770 & 0.760\\
         SOTA \cite{capellan2024model} & 0.659 & 0.646 & 0.595 & 0.496 & \textbf{0.733} & 0.626 & 59.05 & 51.69 & 126.79 & 82.88 & \textbf{99.73} & 84.03 & 0.697 & 0.675 & 0.806 & 0.806\\
         Our Model & \textbf{0.681} & \textbf{0.666} & \textbf{0.652} & \textbf{0.524} & 0.685 & \textbf{0.642} & \textbf{46.00} & 44.23 & \textbf{103.91} & \textbf{69.64} & 100.10 & \textbf{72.96} & 0.709 & \textbf{0.703} & \textbf{0.812} & \textbf{0.813}\\
        \hline
    \end{tabular}
    \label{tab:cbtnScores}
\end{table*}
}

{
\setlength{\tabcolsep}{3pt}
\begin{table*}[h]
    \centering
    \small
    \caption{\small{Lesion-wise Dice Scores, HD95 (mm), Precision and Recall comparing our model against the SOTA in BraTS 2023 GLI testing set.}}
    \begin{tabular}{|c|cccccc|cccccc|cc|cc|}
        \hline
        \multirow{2}{*}{} & \multicolumn{6}{c|}{\textbf{Lesion-wise Dice Scores}} & \multicolumn{6}{c|}{\textbf{Lesion-wise HD95 (mm)}} & \multicolumn{2}{c|}{\textbf{Precision}} & \multicolumn{2}{c|}{\textbf{Recall}}\\
        \cline{2-17}
        Model & WT & TC & ET & NC & ED & AVG & WT & TC & ET & NC & ED & AVG & WT & TC & WT & TC\\
        \hline
         nnU-Net \cite{isensee2021nnu} & 0.869 & \textbf{0.609} & \textbf{0.818} & \textbf{0.375} & \textbf{0.507} & \textbf{0.636} & 21.34 & 36.73 & 14.21 & 55.07 & \textbf{73.80} & \textbf{40.23} & 0.948 & \textbf{0.565} & 0.876 & 0.936\\
         SegMamba 3L/WT & 0.864 & 0.507 & 0.800 & 0.242 & 0.363 & 0.555 & \textbf{21.19} & 35.55 & \textbf{13.27} & 53.89 & 122.36 & 49.25 & 0.882 & 0.400 & \textbf{0.937} & \textbf{0.955}\\
         SOTA \cite{capellan2024model} & 0.840	& 0.447 & 0.720 & 0.200 & 0.000 & 0.441 & 31.14 & 41.75 & 50.32 & 57.64 & 374.00 & 110.97 & \textbf{0.955} & 0.337 & 0.837 & 0.924\\
         Our Model & \textbf{0.877} & 0.549 & 0.815 & 0.286 & 0.449 & 0.595 & 22.43 & \textbf{35.11} & 19.91 & \textbf{53.62} & 103.55 & 46.92 & 0.916 & 0.450 & 0.925 & 0.949\\
        \hline
    \end{tabular}
    \label{tab:gliScores}
\end{table*}
}

\section{Results}
\label{sec:results}

\subsection{BraTS PED Segmentation Results}
\label{ssec:bratsresults}
We evaluated four models' performance on our 26-patient testing set from PED BraTS 2024. Using both the nnU-Net and SegMamba frameworks, we trained one baseline model each on all 4 labels and another model using the 3L/WT segmentation approach for each framework. For both frameworks we see improvements in lesion-wise Dice, lesion-wise HD95, precision and recall scores almost across the board, as seen in Table 1. For SegMamba, we see improvements in all Dice scores, with the average score improving by 1.7 percent and the HD95 distance reducing by 5.89 mm on average. Likewise, when using the nnU-Net framework, there is an improvement in almost all metrics, with the average lesion-wise Dice score improving by 2.2 percent and the lesion-wise HD95 decreasing by 8.04 mm. Recall and precision scores for WT and TC also increase for the respective 3L/WT models compared to their baselines. These are clear performance increases when using the 3L/WT segmentation logic relative to each frameworks baseline which is consistent regardless of backbone. Therefore, the 3L/WT segmentation logic increases segmentation accuracy but also validates the robustness of the segmentation logic since the results were consistent across different backbones.

\subsection{CBTN Segmentation Results}
\label{ssec:cbtnresults}
When testing our model in a real-world out-of-distribution dataset, a decrease in performance of lesion-wise Dice scores and lesion-wise HD95 scores was observed relative to performance on the PED BraTS 2024 testing set. However, when compared to the SOTA on the CBTN dataset, we observed improvements in all lesion-wise metrics except for edema. On average, we saw a rise in average lesion-wise Dice scores of 1.6 percent over the SOTA model and a decrease in average lesion-wise HD95 distances of 11.07 mm as seen in Table 2. Likewise we saw increased performance in recall and precision over the SOTA on both the WT and TC. This represents an improvement in scores on a real-world dataset over the SOTA model in lesion-wise Dice scores, lesion-wise HD95 scores, precision and recall. Some examples of output masks are presented in Fig. 2.

\subsection{BraTS Adult Glioma 2023 Results}
\label{ssec:BraTS Adult Glioma 2023 Results}

In the adult dataset we tested the same 4 models, all trained on pediatric brain tumor segmentation, and saw varying results. When comparing our model against the SOTA, our model is performing better in all metrics barring WT precision, as seen in Table 3. Additionally, our model demonstrates superiority in the WT segmentation with the highest Dice score of 0.877 which 3.7 percent better than SOTA. The baseline nnU-Net did perform well with the best results in the remaining lesion-wise Dice scores. There is a large variety in models with the best performance in precision, recall, and HD95 scores, as seen in Table 3. Overall, our model's average lesion-wise Dice score was 15.4 percent higher than the SOTA and 4.1 percent lower than nnU-Net. Likewise, the average lesion-wise HD95 score of our model was 64.05 mm lower than the SOTA and 6.69 mm higher than nnU-Net.

\section{Concluding Remarks}
\label{sec:discandconc}
In this paper, we introduce a new approach for pediatric brain tumor segmentation, comparing with the current state-of-the-art pediatric brain tumor segmentation method \cite{capellan2024model} on a real-world testing set. Our approach uses two specialized nnU-Net models: one trained exclusively on WT and the other on ET, CC, and ED regions, inferring NET through post-processing. Experimental results demonstrate that our dual-model configuration significantly outperforms a single nnU-Net trained on all four labels, achieving better segmentation performance across both the PED BraTS 2024 test set and the out-of-distribution CBTN dataset.

To further validate our architecture, we applied the same dual-model approach with SegMamba and observed comparable performance gains, underscoring the robustness of our method. When benchmarked against the PED BraTS 2023 winning model \cite{capellan2024model}, our model yielded superior metrics across all regions except ED (comparable). Our framework, limiting complexity by inferring only NET, allows for more accurate spatial learning of ET, CC, and ED relationships, providing improved segmentation without needing the added NET labels.

Our model exhibited competitive performance compared to other algorithms on a structurally distinct dataset from pediatric brain tumors. In the adult BraTS dataset our model outperformed the SOTA in all metrics besides WT precision and was able to most accurately predict the WT mask. However, as a result of the postprocessing, when the model does not segment ET, CC or ED, any part of WT not belonging to these categories will be classified as NET. The pediatric dataset contains far less ED than the adult dataset making it difficult for all tested models to accurately predict ED. This results in the model predicting NET in place of ED since it will classify a lesion as part of the tumor but is unable to segment the ED. In contrast, the baseline nnU-Net will leave this part of the mask empty. For the lesion-wise Dice scores this will result in better performance for nnU-Net on the remaining lesions. Therefore, further training on examples with more ED would aid in model performance within the adult dataset.

Unlike the PED BraTS 2023 SOTA model, which uses ensemble methods to train on WT, TC, and ED labels before post-processing to derive ET and NC (a combination of non-enhancing tumor, cystic component, and necrosis), our model relies on a simpler yet powerful post-processing technique inspired by radiological practices. This approach, focused on preserving radiological accuracy, enhances segmentation precision for each critical subregion, an essential factor in assessing therapy response and predicting patient survival. We believe that this methodology not only advances pediatric brain tumor segmentation but also aligns closely with clinical needs by prioritizing interpretability and segmentation fidelity as seen in our results in BraTS Adult 2023 Glioma dataset in addition to our surpassing performance in pediatric brain tumor.

\section{Compliance with ethical standards}
\label{sec:ethics}
This research study utilized publicly available human subject data from the BraTS Challenge 2024. As certified users, we accessed the data via Synapse ID (syn53708249) and adhered to the ethical guidelines and criteria for utilizing open-access datasets in scientific research.
With IRB approval and a Data User Agreement between the institutions, we accessed CBTN's anonymized data repository.

\section{Acknowledgments}
\label{sec:acknowledgments} This work is supported by Malnati Brain Tumor Initiative, Northwestern University. We would like to express our sincere gratitude to the CBTN for their exceptional data repository and invaluable support for Pediatric Brain Tumor Research. All other authors have no COI (conflict of interest).

\bibliographystyle{IEEEbib}
\bibliography{refs}

\begin{thebibliography}{10}

\bibitem{kazerooni2023brain}
Anahita~Fathi Kazerooni, Nastaran Khalili, Xinyang Liu, Debanjan Haldar, Zhifan Jiang, Syed~Muhammed Anwar, Jake Albrecht, Maruf Adewole, Udunna Anazodo, Hannah Anderson, et~al.,
\newblock ``The brain tumor segmentation (brats) challenge 2023: Focus on pediatrics (cbtn-connect-dipgr-asnr-miccai brats-peds),''
\newblock {\em arXiv preprint arXiv:2305.17033}, 2023.

\bibitem{lilly2023children}
Jena~V Lilly, Jo~Lynne Rokita, Jennifer~L Mason, Tatiana Patton, Stephanie Stefankiewiz, David Higgins, Gerri Trooskin, Carina~A Larouci, Kamnaa Arya, Elizabeth Appert, et~al.,
\newblock ``The children's brain tumor network (cbtn)-accelerating research in pediatric central nervous system tumors through collaboration and open science,''
\newblock {\em Neoplasia}, vol. 35, pp. 100846, 2023.

\bibitem{kazerooni2024brain}
Anahita~Fathi Kazerooni, Nastaran Khalili, Deep Gandhi, Xinyang Liu, Zhifan Jiang, Syed~Muhammed Anwar, Jake Albrecht, Maruf Adewole, Udunna Anazodo, Hannah Anderson, et~al.,
\newblock ``The brain tumor segmentation in pediatrics (brats-peds) challenge: Focus on pediatrics (cbtn-connect-dipgr-asnr-miccai brats-peds),''
\newblock {\em arXiv preprint arXiv:2404.15009}, 2024.

\bibitem{pati2020cancer}
Sarthak Pati, Ashish Singh, Saima Rathore, Aimilia Gastounioti, Mark Bergman, Phuc Ngo, Sung~Min Ha, Dimitrios Bounias, James Minock, Grayson Murphy, et~al.,
\newblock ``The cancer imaging phenomics toolkit (captk): technical overview,''
\newblock in {\em Brainlesion: Glioma, Multiple Sclerosis, Stroke and Traumatic Brain Injuries: 5th International Workshop, BrainLes 2019, Held in Conjunction with MICCAI 2019, Shenzhen, China, October 17, 2019, Revised Selected Papers, Part II 5}. Springer, 2020, pp. 380--394.

\bibitem{davatzikos2018cancer}
Christos Davatzikos, Saima Rathore, Spyridon Bakas, Sarthak Pati, Mark Bergman, Ratheesh Kalarot, Patmaa Sridharan, Aimilia Gastounioti, Nariman Jahani, Eric Cohen, et~al.,
\newblock ``Cancer imaging phenomics toolkit: quantitative imaging analytics for precision diagnostics and predictive modeling of clinical outcome,''
\newblock {\em Journal of medical imaging}, vol. 5, no. 1, pp. 011018--011018, 2018.

\bibitem{vossough2024training}
Arastoo Vossough, Nastaran Khalili, Ariana~M Familiar, Deep Gandhi, Karthik Viswanathan, Wenxin Tu, Debanjan Haldar, Sina Bagheri, Hannah Anderson, Shuvanjan Haldar, et~al.,
\newblock ``Training and comparison of nnu-net and deepmedic methods for autosegmentation of pediatric brain tumors,''
\newblock {\em American Journal of Neuroradiology}, 2024.

\bibitem{baid2021rsna}
Ujjwal Baid, Satyam Ghodasara, Suyash Mohan, Michel Bilello, Evan Calabrese, Errol Colak, Keyvan Farahani, Jayashree Kalpathy-Cramer, Felipe~C Kitamura, Sarthak Pati, et~al.,
\newblock ``The rsna-asnr-miccai brats 2021 benchmark on brain tumor segmentation and radiogenomic classification,''
\newblock {\em arXiv preprint arXiv:2107.02314}, 2021.

\bibitem{xing2024segmamba}
Zhaohu Xing, Tian Ye, Yijun Yang, Guang Liu, and Lei Zhu,
\newblock ``Segmamba: Long-range sequential modeling mamba for 3d medical image segmentation,''
\newblock {\em arXiv preprint arXiv:2401.13560}, 2024.

\bibitem{isensee2021nnu}
Fabian Isensee, Paul~F Jaeger, Simon~AA Kohl, Jens Petersen, and Klaus~H Maier-Hein,
\newblock ``nnu-net: a self-configuring method for deep learning-based biomedical image segmentation,''
\newblock {\em Nature methods}, vol. 18, no. 2, pp. 203--211, 2021.

\bibitem{kazerooni2024brats}
Anahita~Fathi Kazerooni, Nastaran Khalili, Xinyang Liu, Debanjan Haldar, Zhifan Jiang, Anna Zapaishchykova, Julija Pavaine, Lubdha~M Shah, Blaise~V Jones, Nakul Sheth, et~al.,
\newblock ``Brats-peds: Results of the multi-consortium international pediatric brain tumor segmentation challenge 2023,''
\newblock {\em arXiv preprint arXiv:2407.08855}, 2024.

\bibitem{capellan2024model}
Daniel Capell{\'a}n-Mart{\'\i}n, Zhifan Jiang, Abhijeet Parida, Xinyang Liu, Van Lam, Hareem Nisar, Austin Tapp, Sarah Elsharkawi, Maria~J Ledesma-Carbayo, Syed~Muhammad Anwar, et~al.,
\newblock ``Model ensemble for brain tumor segmentation in magnetic resonance imaging,''
\newblock {\em arXiv preprint arXiv:2409.08232}, 2024.

\end{thebibliography}

\end{document}